\documentclass[letterpaper]{article}
\usepackage{aaai}
\usepackage{times}
\usepackage{helvet}
\usepackage{courier}
\usepackage{graphicx}
\graphicspath{ {./} }
\usepackage{lineno}

\nocopyright

\begin{document}
%
\title{Inverse mapping of face GANs}

\author{Nicky Bayat \\ Western University \\ London, ON, Canada \\ nbayat5@uwo.ca \And Vahid Reza Khazaie \\ Western University \\ London, ON, Canada \\ vkhazaie@uwo.ca \And
Yalda Mohsenzadeh \\ Western University \\ London, ON, Canada \\ ymohsenz@uwo.ca}

\maketitle
\begin{abstract}
\begin{quote}
Generative adversarial networks (GANs) synthesize realistic images from a random latent vector. While many studies have explored various training configurations and architectures for GANs, the problem of inverting a generative model to extract latent vectors of given input images has been inadequately investigated. Although there is exactly one generated image per given random vector, the mapping from an image to its recovered latent vector can have more than one solution. We train a ResNet architecture to recover a latent vector for a given face that can be used to generate a face nearly identical to the target. We use a perceptual loss to embed face details in the recovered latent vector while maintaining visual quality using a pixel loss. The vast majority of studies on latent vector recovery perform well only on generated images, we argue that our method can be used to determine a mapping between real human faces and latent-space vectors that contain most of the important face style details. In addition, our proposed method projects generated faces to their latent-space with high fidelity and speed. At last, we demonstrate the performance of our approach on both real and generated faces. 
\end{quote}
\end{abstract}

\section{Introduction}
Generative adversarial networks \cite{goodfellow2014generative} train two networks simultaneously, a generator and a discriminator. The generator network aims at synthesizing samples as realistic as possible while the discriminator's objective is to distinguish between real and fake samples. This min-max two player game optimizes both networks to compete with each other and boost their respective performances. When given a training dataset, GANs are capable of generating new samples that have similar characteristics with the training samples. Generators receive random latent vectors that are usually sampled from either uniform or normal distributions.

We can invert the generator to recover the latent vector of a given image. Recovered latent vectors can be used as a measure of GAN performance \cite{creswell2018inverting} as well as a method to find out about the features a GAN has learned from its training dataset. In the case that a proper latent vector is not found for an image, we can conclude that certain features in the image cannot be modeled by the generator. This conclusion can be used as a quantitative measure to compare performances of different GANs. Moreover, by extracting proper latent vectors of real images, we can train GANs to modify images of the natural image manifold towards a desired direction \cite{zhu2016generative}\cite{huh2020transforming}, for example applying styles to real faces. In addition, linear operations on latent vectors result in meaningful changes in generated images \cite{radford2015unsupervised}, for example adding latent vectors of a person who is smiling to the latent vector of a neutral face will result in a new smiley face. Another example is linearly transforming the latent vector to manipulate memorability of generated images \cite{goetschalckx2019ganalyze}. Thus mapping from image space to latent space can be useful in classification or retrieval tasks.

Despite the many applications for inverting generators for real images, there are only a few papers focusing on this topic. The majority of related publications focus on retrieving latent vectors of generated images. Recent latent-vector-recovery methods usually perform poorly on real images and even those that have demonstrated results on real image datasets are rarely tested on human faces. Therefore, we believe a novel approach that is capable of regaining the z-space vector of a given natural image will be an asset to a variety of applications.

\cite{creswell2018inverting} proposed an approach to invert any pre-trained GAN using gradient descent with pixel-loss as the reconstruction loss. This method recovers the latent vector of an input image in a way that when fed into the GAN a similar image to the target is generated. They indicate that this inverted GAN can be used as both a qualitative and quantitative measure for evaluating GANs. The generated faces used in this work have very low quality, here we use progressively growing GANs \cite{karras2017progressive} to synthesize more reliable faces with higher resolution. Additionally, \citeauthor{creswell2018inverting} evaluate their model only on the same dataset with which they trained the GANs, which might not provide information on the method’s performance on other natural images.  The biggest downside of gradient-based methods is that they tend to fall into local minima which results in poor reconstructed latent vectors. Stochastic clipping, introduced in \cite{lipton2017precise}, can recover the true latent vectors for DCGAN \cite{radford2015unsupervised} generated images with high fidelity. Nevertheless, these techniques need to converge the gradient descent for each test image separately. For example \cite{lipton2017precise} needs 20k iterations of gradient descent to reconstruct good images and this can be extremely time consuming especially when we plan to extract latent vectors of many images. Furthermore, recent GANs are becoming much deeper and are much more challenging to invert than DCGAN \cite{bau2019seeing}.

Another common approach for extracting latent space vectors of images is training an encoder network alongside the GAN \cite{donahue2016adversarial}\cite{dumoulin2016adversarially}. The encoder learns to invert the generated image to its original latent vector. The downside of this technique is that the encoder might overfit and lead to poorly reconstructed images, specifically those images that are drawn from a different distribution than the training set. In addition, this method is not applicable for pre-trained GANs because it has to be trained with the GAN at the same time. Moreover, training a third network adds more parameters and therefore is not efficient in resource consumption. The method described in \cite{luo2017learning} learns an encoder but after training the GAN. Hence, this method is able to handle pre-trained networks. However, the problem of overfitting persists.

In this paper, our focus is on inverse mapping of face GANs. We train a residual neural network (ResNet18) \cite{he2016deep} to extract latent vectors of given face images. This study focuses on recovering latent vectors of both generated faces in addition to extracting face details and styles of real human faces. First, we train ResNet with only generated faces using pixel and perceptual loss. This approach performs well in the inverse mapping of generated faces and finds nearly identical recovered latent vectors to the ground truths three orders of magnitude faster than gradient-based alternatives.

Transferring styles of images using GANs has been the subject of many recent studies. In this investigation, we focus on extracting face details such as hair style, gender, pose, facial expressions, etc. when extracting latent vectors of the natural faces. Then we use the recovered vector to generate new faces that contain the same styles as the real input face images. In order to map natural faces to latent space vectors that contain their style information, we train the network simultaneously on generated and natural face images. We transfer latent-space information from generated faces while extracting detailed facial information using pixel and perceptual loss of the real faces. Training based on pixel loss between recovered latent vectors of generated images and their ground truth latent vectors (z-loss) helps the model to learn the mapping between image-space and latent-space. The pixel and perceptual loss between reconstructed and natural images aids the network to generate faces with similar attributes to the target.

In this work our contributions are the following:
\begin{itemize}
\item We propose a new approach for recovering latent vectors of generated faces using deep residual neural networks. This technique reconstructs images similar to the target much faster than optimization-based solutions.
\item Using a combination of pixel, perceptual, and z-loss, we are able to map given natural face images to latent space counterparts that contain similar facial information and style attributes. The reconstructed latent vector can be fed to a GAN to generate an image with identical characteristics to the target.
\item We present results of latent vector recovery for images from other datasets (that GAN was not trained on) which proves the generalizability of our model to other datasets.
\end{itemize}

\begin{figure*}[t]
\centerline{\includegraphics{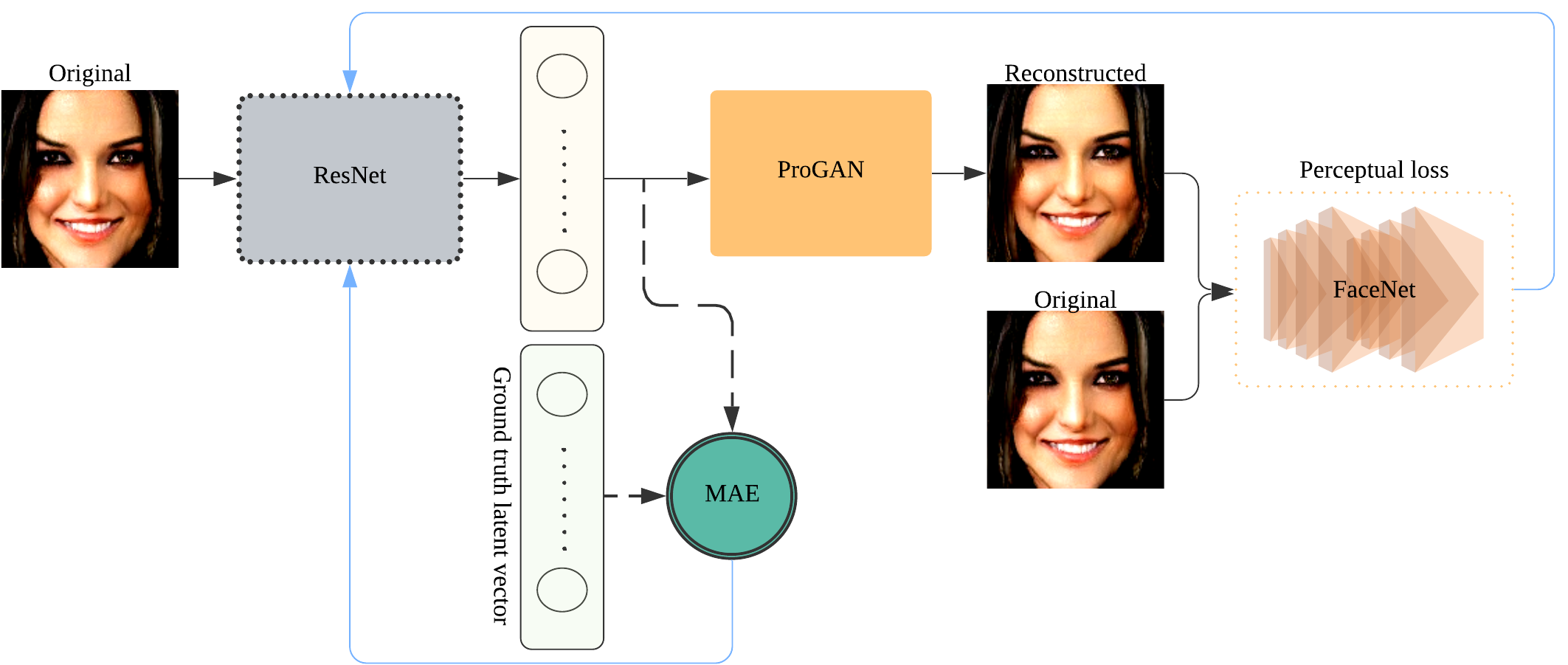}}
\caption{The proposed framework for mapping generated faces to latent-space vectors.}
\label{fig1}
\end{figure*}

\section{Related Works}
Despite the many applications of GAN inversion, this area is still an open research problem when dealing with natural input images, specifically pictures of human faces. Methods focusing on projecting images into the corresponding GAN latent vectors can be categorized into four major groups. 

The first category trains an encoder network alongside the GAN \cite{donahue2016adversarial}\cite{dumoulin2016adversarially}. The encoder learns to map the generated image back to its latent vector. The encoder adds more parameters to the training, which is not efficient in terms of resource usage. Moreover, this network can easily overfit to the training data, which leads to poor reconstructed latent vectors as well as substandard performance as an evaluation measure for GANs. In addition, such techniques require simultaneous training with the GAN, meaning they cannot be used on pre-trained GANs. Such drawbacks have motivated more recent studies to move in a different direction. \cite{luo2017learning} have trained the encoder based on a pre-trained GAN. Although this method overcomes the challenge of handling pre-trained networks, the problems of overfitting and inefficient resource consumption persist.

A large and growing body of the literature has investigated optimization-based solutions \cite{creswell2018inverting}\cite{lipton2017precise}. They use gradient descent to optimize the problem of projecting images into the latent vectors that can be used to regenerate those images. They start with a random latent vector, z', sampled from the distribution on which the generator was trained. The goal is to minimize the L2 norm between the target image and the image generated by z' using gradient descent. At each iteration, z' is updated to generate an image more accurate to the target until they converge. \citeauthor{lipton2017precise} introduced stochastic clipping to bound reconstructed latent vectors to the original domain. This technique performs better than solely gradient descent with exceptionally high recovery accuracy on DCGAN generated images, which results in generating images indistinguishable from the target. While the results of this work seem promising, they are usually obtained after numerous iterations of gradient descent, and therefore time to convergence can be a huge barrier when recovering latent vectors of many images. In addition, more recent deep GANs, such as the 15-layer progressive GAN used in this work, are much more challenging to invert than a DCGAN. Additionally, this approach performs poorly when applied to real input faces.

The third category of methods combines the previously mentioned approaches to benefit from both. These methods first use an encoder to obtain initial z', then use gradient descent to optimize the problem \cite{bau2020semantic}\cite{bau2019seeing}. Despite the fact that recovered vectors are very similar to the original z-space vector, they generate blurry images with inadequate texture details.

\begin{figure*}[t]
\centerline{\includegraphics{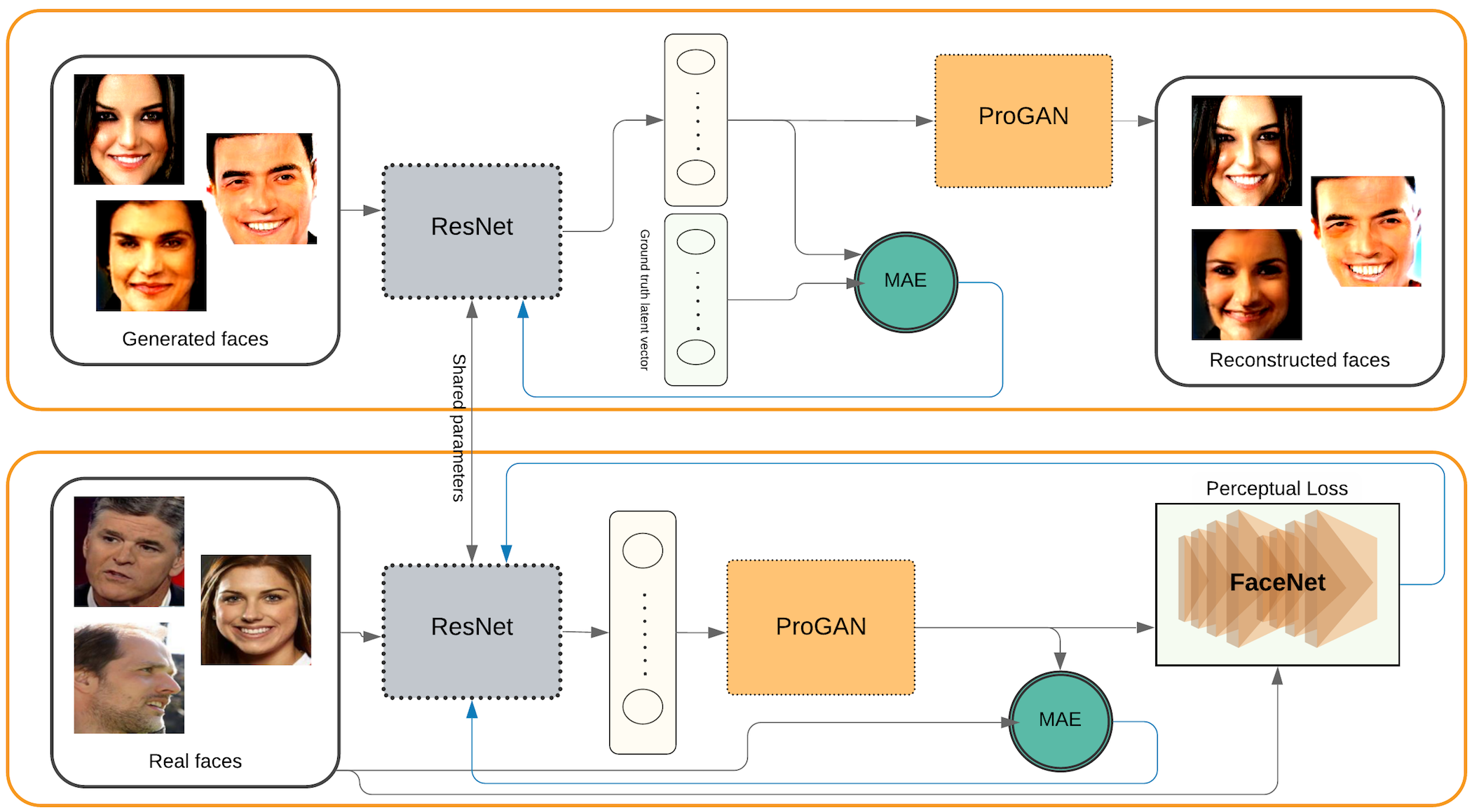}}
\caption{The proposed framework for mapping natural faces to latent-space vectors.}
\label{fig2}
\end{figure*}

Recent work proposed in \cite{huh2020transforming} has shown improvement in transforming and recovering latent vectors of natural images. They use a gradient-free Covariance Matrix Adaptation (CMA) optimizer to recover the corresponding latent vector of a given target image. They have also evaluated various optimizers for the GAN inversion problem and found BasinCMA \cite{wampler2009optimal} to be the best. This technique updates CMA after taking a predefined number of gradient-descent steps, which benefits from the advantages of both methods. Despite outstanding improvement in visual quality of the generated images using recovered latent vectors, this work focused on BigGAN \cite{brock2018large} pre-trained on ImageNet \cite{deng2009imagenet}, which does not have a human face class. Moreover, this approach still requires a considerable amount of time to converge for each input image.

A number of studies have begun to examine neural-network-based approaches with perceptual similarity metrics to map images to latent-space \cite{zhu2016generative}\cite{dosovitskiy2016generating}. The neural network uses a combination of pixel and perceptual loss to find the inverse mapping of image-space into latent-space. These methods are much faster but previously were not visually as good as optimization-based techniques. We believe that by using this approach with the right deep neural network and use of proper objective functions, we will be able to achieve faces identical to the target with much less computation time.

\textbf{Style transfer:} Several studies \cite{zhu2016generative}\cite{bau2020semantic}\cite{brock2016neural} have used latent vector recovery methods for real image manipulation and style transfer. They first extract a latent vector that generates an image as similar as possible to the input image, then update the latent representation in order to smoothly transform the generated image and add the desired styles. This transformation is extremely sensitive because it has to apply edits to the image while keeping it inside the natural image manifold. There is a relatively small body of literature that investigates adding styles to natural faces or adding human face styles to generated faces. Even though StyleGAN \cite{karras2019style} focuses on transferring face styles using GANs, its creators use latent z vectors of generated faces to embed styles into different layers of the generator rather than natural images. We argue that our framework is capable of extracting important facial details and embedding them into the recovered latent vector, which will be fed into a pre-trained generator to synthesis the desired output. Hence, our approach can be used to add natural-face styles to generated images of any pre-trained generator.

\section{Method}
\subsection{Overview}
In this work, we train a residual neural network (ResNet18) in order to map an input image to its corresponding latent vector using a combination of a reconstruction loss and a perceptual loss. Mean Absolute Error (MAE) is used as the reconstruction loss.

We introduce two frameworks: the first architecture (Figure 1) trains the network on generated faces for which we have the ground truth latent vectors. The second architecture deals with natural human faces using a pixel loss and a perceptual loss between the reconstructed face and the target as well as the z-loss. The overview of our proposed framework for real human faces is depicted in Figure 2.
\subsection{Generative adversarial networks}
GANs train two networks simultaneously. The generator network aims at synthesizing fake samples as realistic as possible while the discriminator's goal is to distinguish between real and fake samples. This min-max two player game and the competition between the generator and the discriminator optimizes their performance. Given a training dataset, GANs are capable of generating new samples that have similar characteristics with the training samples.

\subsection{Mapping strategy}
Our goal is to train a deep neural network that is capable of mapping samples from image-space into latent vector space (z-space) using a dataset of generated images and their matching ground truth latent vectors. Since residual blocks are a necessity when it comes to training deep networks \cite{zhang2017beyond} we decided to train a residual neural network (ResNet18) to find the mapping. A training dataset of 100k generated faces and their ground truth latent vectors was created using a progressive GAN (proGAN) trained on 128x128 CelebA faces \cite{liu2015deep}.

When training on only generated faces, MAE between features extracted from the last layer of ResNet18 and ground truth latent vectors, as well as the perceptual loss between reconstructed and target images, were back propagated through the network. Our proposed method (Figure 1) is capable of recovering latent vectors similar to the ground truths for given validation images with high fidelity and speed.

Next, we attempted to find the mapping for natural human faces by using MAE and perceptual loss between the reconstructed image and target as well as simultaneous training of ResNet on both natural and generated faces. The combination of MAE and perceptual loss helped the model to generate images similar to the real target, while training on generated faces with their ground truth latent vectors helped ResNet in learning an accurate inverse mapping.

\begin{figure*}[!htp]
\centerline{\includegraphics{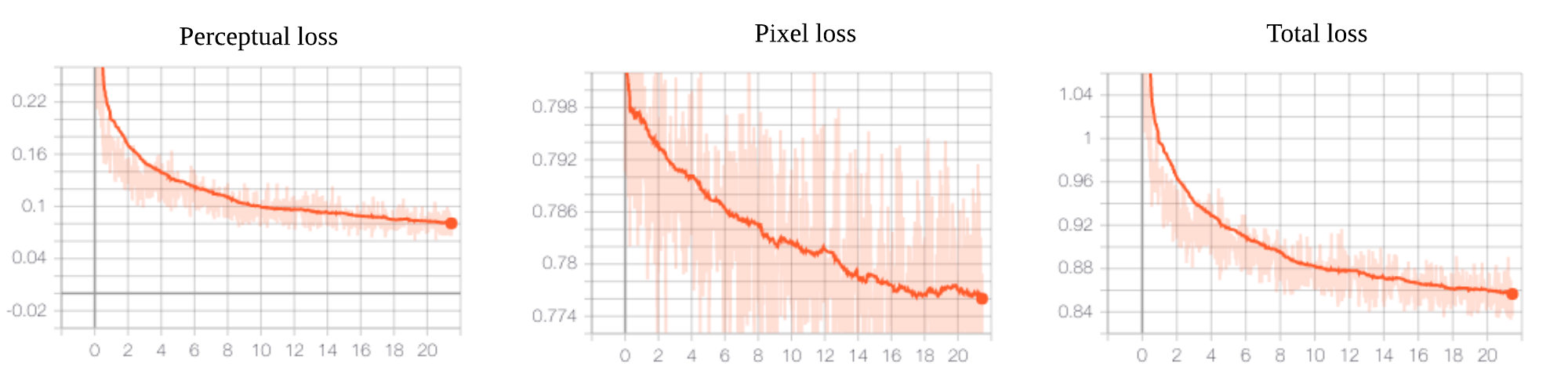}}
\caption{The loss when training ResNet on generated faces. Vertical and horizontal axes show the loss and hours of training respectively.}
\label{fig3}
\end{figure*}

\subsection{Objective function}
\textbf{Generated images:} When training ResNet on generated faces, we were able to use ground truth latent vectors, random vectors that were originally used to generate training images. The MAE loss between features extracted from the last layer of ResNet and the corresponding ground truth latent vector, the z-loss, was fed backwards into the network in order to find the best inverse mapping.

Recent work on perceptual loss for style transfer \cite{johnson2016perceptual} shows that we are able to generate images with high fidelity by defining and optimizing a perceptual loss. The perceptual loss can be extracted from various layers of pre-trained networks. We obtained perceptual loss from 21 concatenation layers of a FaceNet \cite{schroff2015facenet} pre-trained on MS-Celeb-1M. Adding perceptual loss results in visually better faces with adequate texture.

\textbf{Natural images:} Since the ground truth latent vectors for real faces are not available, we used a reconstruction loss which is the MAE between real faces and generated faces using features extracted from ResNet. We added also perceptual loss to embed texture details and improve visual quality. The average of the Mean Square Error (MSE) loss between features extracted from all concatenation layers of FaceNet for real and reconstructed images determines the perceptual loss at each epoch. If we only train ResNet on pixel and perceptual loss of real faces, the model forgets the mapping of image-space into latent-space. In order to overcome this challenge, we trained ResNet one epoch on z-loss of generated faces and one epoch on pixel and perceptual loss of real faces.
\begin{figure}
\centerline{\includegraphics{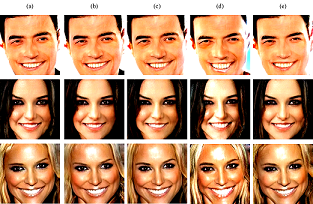}}
\caption{Original generated faces are presented in column (a). Column (b) is the generated faces of recovered latent vectors by (a) using gradient-descent method with 200 iterations. Column (c) applies stochastic clipping while updating gradient descent. Column (d) is our method, which utilizes our trained ResNet to map generated images to their corresponding latent vectors. Column (e) is our ResNet trained using both pixel and perceptual loss.}
\label{fig4}
\end{figure}
\begin{figure*}
\centerline{\includegraphics{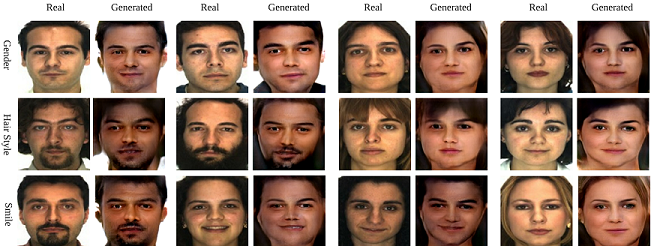}}
\caption{The results for mapping natural faces to latent-space vectors that contain same style and facial features. The faces in each row represent how recovered latent vectors understand the gender, hair style and emotions of the target image respectively.}
\centering
\label{fig5}
\centerline{\includegraphics{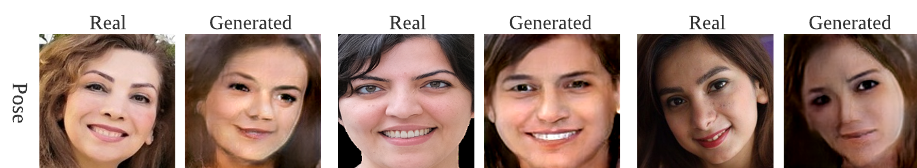}}
\caption{Recovered latent vectors preserve the pose of the target face.}
\label{fig6}
\end{figure*}
\subsection{Training and implementation details}
This work was implemented in Tensorflow 2. Three fully connected layers were added to the ResNet18 architecture with 2048, 1024 and 512 units respectively. ResNet was trained using an Adam optimizer with a 2e-4 learning rate on a combination of pixel and perceptual loss. The training loss over time (per hour) is shown in Figure 3. The progressive GAN, which is used to generate faces, was pre-trained on 128x128 CelebA faces and was downloaded from Tensorflow Hub. A FaceNet pre-trained on the MS-Celeb-1M network was used to compute perceptual loss.

\textbf{Datasets:} 100k faces were generated with proGAN using random latent vectors sampled from normal distribution to create our generated faces training set. In order to obtain our real faces dataset, we first selected all the VggFace2 \cite{cao2018vggface2} identities that had at least 100 faces, out of which we randomly selected 100 samples per identity and then randomly picked 100 identities. Hence, our final real human face dataset consisted of 10,000 images.
\begin{table}[htbp]
\caption{Comparing gradient-based methods with ours based on PSNR, FID score and the computation time (in seconds) for each method.}
\begin{center}
\begin{tabular}{|c|c|c|c|}
\hline
\textbf{}&\multicolumn{3}{|c|}{\textbf{Metric}} \\
\cline{2-4}
\textbf{Model} & \textit{PSNR}& \textit{FID}& \textit{Time} \\
\hline
Adam&18.41&1.96&1143.90\\
Stochastic Clipping&\textbf{19.06}&1.98&1135.24\\
Pixel-ResNet(ours)&11.06&\textbf{1.94}&\textbf{3.27}\\
Pixel-Perceptual ResNet(ours)&16.30&1.98 &3.96\\
\hline
\end{tabular}
\label{tab1}
\end{center}
\end{table}
\section{Experiments and Results}
\subsection{Latent vector recovery of generated faces}
In order to evaluate the performance of our model in mapping generated faces to latent vector peers, we set up an experiment to both qualitatively and quantitatively compare our results with gradient-based methods. 

Gradient-based techniques update gradient descent for a variable number of iterations, so choosing the right number of iterations is critical in obtaining optimal results. In some cases, updating is stopped when the loss is below a certain threshold, which means a different number of iterations is required for different generated faces and some might need up to 100k iterations of gradient descent to converge. In this work we chose 200 iterations to make our comparison viable in terms of computation time. 

We generated 50 faces using proGAN to test the recovery of latent vectors using four different methods: gradient descent, gradient descent with stochastic clipping, our ResNet model trained based on pixel loss, and our ResNet trained on both pixel and perceptual loss. Gradient descent was implemented using an Adam optimizer with 0.01 learning rate and MAE loss. Stochastic clipping binds latent vectors to [-1, 1] range. If a number is less than -1 or bigger than 1, we simply replace it with a random number within the range. Our ResNet18 architecture for this experiment was trained using the framework shown in Figure 1. The experiment was conducted on a Nvidia GeForce RTX 2080 TI GPU.

Qualitative results are presented in Figure 4. We compared our method trained with solely pixel loss and also both pixel and perceptual loss with gradient-based alternatives. Adding perceptual loss improves visual quality and results in faces indistinguishable from the target (see the comparison of Figure 4a (targets) and Figure 4e (generated images with recovered latent vectors using our method).

Quantitative results are reported in Table 1. What stands out in the table is that our method is significantly faster than other state-of-the-art techniques while having slightly lower PSNR. We compromise visual quality (to a negligible degree) for three orders of magnitude faster latent vector recovery. The difference in PSNR metric is negligible compared to the significantly lower computation time. Interestingly, in terms of Frechet Inception Distance (FID) score, which is calculated based on cosine distance between embeddings extracted from the last layer of the pre-trained FaceNet, our method performs slightly better than gradient-based alternatives. We can therefore conclude that our recovered latent vectors reconstruct better face features and also perform better in identification tasks.

\subsection {Style transfer using natural faces}
Mapping real human faces to latent-space vectors, which leads to regenerating those faces using GANs, is extremely challenging and very little was found in the literature on this topic. Our proposed architecture for natural faces extracts latent vectors that contain important facial information on the given real face. These facial details include face shape, gender, hair/beard style, smile, pose, etc. and will then be applied to the generated faces. This approach can be a means to control the face image synthesis process by altering the latent vector based on given styles.

In order to evaluate our ResNet model in style transformation from real to generated faces, we conducted an experiment on the AR face dataset \cite{martinez1998ar}. AR is a strictly constrained face dataset that includes over 4000 frontal view face images of 126 identities (70 males and 56 females). Facial impressions, illumination and occlusion of the participants are controlled, but there are no restrictions on what people are wearing (make up, glasses, hair style, etc.). The pictures were taken in the same conditions in two different sessions separated by two weeks.

The findings of this experiment indicate that projecting a real face image into latent-space using a deep residual network trained on pixel, perceptual and z-loss results in a latent vector that can be used to generate a face with similar styles as the input.  Figure 5 illustrates the style transformation of a generated face given natural input faces on the AR face dataset. The recovered latent vector includes information about the gender, hair/beard style, facial expressions (for example smile) and many other attributes of the natural face.

In addition, we conducted another experiment on high quality pictures of real human faces with different poses to indicate that recovered latent vectors not only preserve hair style, gender and facial expressions, but also generate faces with the same pose (Figure 6).

\section{Conclusion}
In this paper, we proposed a method to project generated face images into latent-space extremely fast with high visual quality using deep residual neural networks trained with pixel and perceptual loss. The second major finding was that training the same architecture on pixel and perceptual loss for real faces plus z-loss between generated faces and their ground truth latent vectors will bring about a latent vector that contains important facial details and styles of the input face, such as gender, pose, emotion, hair/beard style,etc.

The results of this research support the idea that, even though optimization-based approaches are performing well in inverting generators and obtaining accurate latent vectors, deep residual networks in our method are capable of performing the same tasks with relatively identical quality and incredibly faster speed.

The insights gained from this study may be of assistance in other problems that involve mapping a huge number of images to their corresponding latent vectors and then finding a new mapping or transformation within the latent-space to generate desired images.


In this study, we substantiated that training a deep neural network on a combination of pixel, perceptual and z-loss can be used to create a latent vector that includes important details and styles of the target real human faces. Though, it still remains an open question if recovering an accurate latent vector for real faces is possible, especially in terms of identity information.

\bibliographystyle{aaai}\bibliography{Bayat}
\end{document}